# Training the next generation of physicians for artificial intelligence - assisted clinical neuroradiology:

# ASNR MICCAI Brain Tumor Segmentation (BraTS) 2025 Lighthouse Challenge education platform on neuroradiology and artificial intelligence

Raisa Amiruddin[1]*, Nikolay Y. Yordanov[2]*, Nazanin Maleki[1], Pascal Fehringer[3], Athanasios Gkampenis[4], Anastasia Janas[5], Kiril Krantchev[5], Ahmed Moawad[6], Fabian Umeh[7], Salma Abosabie[8], Sara Abosabie[5], Albara Alotaibi[9], Mohamed Ghonim[10], Mohanad Ghonim[11], Sedra Abou Ali Mhana[6], Nathan Page[3], Marko Jakovljevic[1], Yasaman Sharifi[12], Prisha Bhatia[13], Amirreza Manteghinejad[1], Melisa Guelen[5,14], Michael Veronesi[15], Virginia Hill[16], Tiffany So[17], Mark Krycia[18], Bojan Petrovic[19], Fatima Memon[18,20,21], Justin Cramer[22], Elizabeth Schrickel[23], Vilma Kosovic[24], Lorenna Vidal[1], Gerard Thompson[25,26], Ichiro Ikuta[22], Basimah Albalooshy[1], Ali Nabavizadeh[1,6], Nourelhoda Tahon[27], Karuna Shekdar[1], Aashim Bhatia[1], Claudia Kirsch[28], Gennaro D'Anna[29], Philipp Lohmann[30,31], Amal Saleh Nour[32], Andriy Myronenko[33], Adam Goldman-Yassen[34], Janet R. Reid[1], Sanjay Aneja[35], Spyridon Bakas[36,37,38], Mariam Aboian[1]

*Equal first author contribution

1. Department of Radiology, Children's Hospital of Philadelphia, Philadelphia, PA, USA

2. Department of Neurointensive Care, Multiprofile Hospital for Active Treatment in Neurology and Psychiatry 'St. Naum', Sofia, Bulgaria

3. Faculty of Medicine, Jena University Hospital, Friedrich Schiller University Jena, Jena, Germany

4. Department of Neurosurgery and Neurotechnology, University of Tubingen, Tubingen, Germany

5. Charité – Universitätsmedizin Berlin, Berlin, Germany

6. Department of Radiology, Hospital of the University of Pennsylvania, Philadelphia, PA, USA

7. School of Computing, Engineering and Digital Technologies, Teesside University, Middlesbrough, North Yorkshire, United Kingdom

8. Institute of Experimental Biomedicine, University Hospital of Wuerzburg, Wuerzburg, Bavaria, Germany

9. Jordan University of Science and Technology, Irbid, Jordan

10. Department of Radiology, University of Mississippi Medical Center, Jackson, MS, USA

11. Department of Radiology, University of Arkansas for Medical Sciences, Little Rock, AR, USA

12. Department of Radiology, Iran University School of Medicine, Tehran, Iran

13. Mohammed Bin Rashid University of Medicine and Health Sciences, Dubai, United Arab Emirates

14. Athinoula A. Martinos Center for Biomedical Imaging, Massachusetts General Hospital, Boston, MA, USA

15. Department of Radiology, University of Wisconsin School of Medicine and Public Health, Madison, WI, USA

16. Department of Radiology, Northwestern University, Feinberg School of Medicine, Chicago, IL, USA

17. Department of Imaging and Interventional Radiology, Chinese University of Hong Kong, Hong Kong

18. Carolina Radiology Associates, Myrtle Beach, SC, USA

19. NorthShore Endeavor Health, Evanston, IL, USA

20. McLeod Regional Medical Center, Florence, SC, USA

21. Medical University of South Carolina, Charleston, SC, USA

22. Department of Radiology, Mayo Clinic, Phoenix, AZ, USA

23. Department of Radiology, Ohio State University College of Medicine, Columbus, OH, USA

24. Department of Radiology, General Hospital of Dubrovnik, Dubrovnik, Croatia

25. Centre for Clinical Brain Sciences, University of Edinburgh, Edinburgh, United Kingdom

26. Department of Clinical Neurosciences, NHS Lothian, Edinburgh, United Kingdom

27. Department of Radiology, University of Missouri, Columbia, MO, USA

28. Department of Radiology and Biomedical Imaging, Yale School of Medicine, New Haven, CT, USA

29. Department of Radiology, Centro Diagnostico Italiano, Milan, Italy

30. Institute of Neuroscience and Medicine, Forschungszentrum Julich, Julich, Germany

31. Department of Nuclear Medicine, Rheinisch-Westfalische Technische Hochschule Aachen University Hospital, Aachen, Germany

32. Department of Radiology, College of Health Science, Addis Ababa University, Addis Ababa, Ethiopia

33. NVIDIA, Santa Clara, CA, USA

34. Children's Healthcare of Atlanta, Emory University School of Medicine, GA, USA

35. Department of Therapeutic Radiology, Yale School of Medicine, New Haven, CT, USA

36. Division of Computational Pathology, Department of Pathology and Laboratory Medicine, Indiana University School of Medicine, Indianapolis, IN, USA

37. Department of Radiology and Imaging Sciences, Indiana University, Indianapolis, IN, USA

38. Department of Neurological Surgery, School of Medicine, Indiana University, Indianapolis, IN, USA

# ABSTRACT

**BACKGROUND**: High-quality reference standard image data creation by neuroradiology experts for automated clinical tools can be a powerful tool for neuroradiology and artificial intelligence education. We developed a multimodal educational approach for students and trainees during the MICCAI Brain Tumor Segmentation Lighthouse Challenge 2025, a landmark initiative to develop accurate brain tumor segmentation algorithms.

**MATERIALS & METHODS**: Fifty-six annotators (medical students & radiology trainees) volunteered to annotate brain tumor MR images for the BraTS challenges of 2023 and 2024, guided by faculty-led didactics on neuropathology MRI. Separately, 54 medical students were surveyed on AI education at their institutions to confirm interest in learning opportunities. Among the 56 annotators, 14 select volunteers were then paired with board-certified neuroradiology faculty for guided one-on-one annotation sessions for BraTS 2025, focusing on brain metastases, meningiomas, and glioblastomas. To expand our reach, lectures on neuroanatomy, pathology and AI, journal clubs, and data scientist-led workshops were organized online. Annotators and audience members completed surveys on their perceived knowledge before and after annotations and lectures respectively.

**RESULTS**: Among 54 medical students, while 93% (50/54) believed that AI would influence their careers, 87% (47/54) reported no AI-focused education at their institutions and 72% (39/54) considered AI focused training to be important. Fourteen coordinators, each paired with a neuroradiologist, completed the data annotation process, averaging 1322.9±760.7 hours per dataset per pair and 1200 segmentations in total. On a scale of 1-10, annotation coordinators reported significant increase in familiarity with image segmentation software pre- and post-annotation, moving from initial average of 6 ± 2.9 to final average of 8.9 ± 1.1 ($p<0.05$), and significant increase in familiarity with brain tumor features pre- and post-annotation, moving from initial average of 6.2 ± 2.4 to final average of 8.1 ± 1.2 ($p<0.05$). Among 97 lecture attendees who completed pre- and post-lecture surveys, 95% (92/97) agreed that their knowledge of topics improved after the lectures.

**CONCLUSION**: We demonstrate an innovative offering for providing neuroradiology & AI education through an image segmentation challenge to enhance trainees' understanding of algorithm development, reinforce the concept of data reference standard, and diversify opportunities for AI-driven image analysis among future physicians.

**ABBREVIATIONS**:

AI = artificial intelligence

ASNR = American Society of Neuroradiology

BraTS = Brain Tumor Segmentation Challenge

MICCAI = Medical Image Computing and Computer Assisted Interventions Society

MRI = magnetic resonance imaging

## SUMMARY SECTION

**PREVIOUS LITERATURE:** The expanding role of AI in radiology necessitates focused training to equip future radiologists with skills to effectively navigate the applications of automated tools in clinical practice. Existing literature calls for medical education reform to incorporate AI into curricula, emphasizing the importance of interdisciplinary collaboration and insight to evaluate the limitations of AI. The Brain Tumor Segmentation (BraTS) Challenge is an established initiative for AI algorithm development using robust annotated data but has demonstrated the critical need for high quality annotated datasets[1–5]. This article details the creation of a novel, interactive learning opportunity leveraging the various steps of the BraTS challenge to build trainee knowledge of AI in brain tumor imaging while simultaneously generating high quality annotated datasets for training of segmentation algorithms.

**KEY FINDINGS:** A hybrid educational approach combining didactic lectures with interactive learning through guided data annotation improved trainees' knowledge with automated image analysis and offered insights for incorporating AI-focused education in medical and radiology curricula to prepare future radiologists for an AI-integrated clinical environment.

**KNOWLEDGE ADVANCEMENT:** This article showcases an evolving educational initiative that couples generation of high-quality annotated imaging datasets with radiology education to provide trainees with interactive learning opportunities to improve AI-related knowledge and technical skills, reinforcing best practices of adult learning. We further propose strategies for medical educators to develop formal AI-focused education in medical school and residency.

## INTRODUCTION

The practice of radiology has fast evolved from the age of information – wherein volumes of information on the internet had to be assimilated by users, to the age of artificial intelligence (AI) - wherein this information is collated, interpreted, and translated by machines to deliver more easily digestible and focused results [6,7]. To effectively leverage developments in AI, radiologists need to possess the skills to analyze and critique data to improve the performance and correctly interpret the automated outcomes [8]. Modern radiology curricula must bridge the gap between theory and practice to equip trainees to function in AI-supported environments as evidenced by the American Board of Radiology inclusion of AI in their board examination content [9]. Hence, there is a critical need to augment training with AI-focused knowledge and skills, which may address the need to develop automated tools to support clinical practice [7]. A significant constraint to addressing this issue is the challenge of recruiting radiology faculty to participate in data annotation and AI-focused curriculum development [10]. We aimed to establish a novel blueprint to streamline data annotation and concurrently provide learning opportunities for trainees to build their knowledge of AI in imaging.

The American Medical Association noted numerous calls for reform in medical education over the past two decades. Medical educators are increasingly motivated to design educational initiatives that focus on the increasing applications of AI in practice, recognizing that the future of patient care lies in collaboration among physicians and AI-based systems [9]. To uphold integrity in delivering healthcare, AI integration necessitates a balanced approach combining technological expertise with ethical considerations, best addressed by interdisciplinary collaboration among physicians, data scientists, and engineers [7].

Learners need exposure to AI in clinical practice to understand the volume, variety, and quality of available data [11]. They also need to understand how AI is used to personalize healthcare through applications like decision support software [12]. This requires systemic reform in radiology education to move beyond traditional models and prepare trainees to critically evaluate the limitations of AI and address ethical, legal, and regulatory issues to ensure patient safety and privacy [13]. Learning is enhanced through exploration, examination, and analysis of information [14]. Interactive learning fosters engagement, guiding learners to apply knowledge in real-world scenarios, and to develop critical thinking and problem-solving skills; this is one of the tenets of adult learning [15,16].

The Brain Tumor Segmentation (BraTS) Challenge is a global annual initiative championed by the American Society of Neuroradiology (ASNR) and Medical Image Computing and Computer Assisted Interventions Society (MICCAI) involving organized collaboration between neuroradiologists and data scientists to train and develop algorithms for automated analysis of brain tumors on MRI, including high-integrity annotated data, considered 'ground truth' or 'reference standard' [17,18]. The BraTS 2025 Challenge organizers created learning opportunities for medical students and radiology trainees to develop their skills and understanding of automated tools in brain tumor imaging. Through engagement with real-world AI challenges, this initiative provided trainees and students with hands-on experience to bridge theory and practice, while emphasizing the importance of accessible and reliable advancements in medical imaging AI.

## MATERIALS AND METHODS

BraTS Challenges of 2023 and 2024:

During the ASNR MICCAI BraTS Challenges of 2023 and 2024, the organizing team incorporated educational activities for medical students and radiology trainees, including residents and research fellows, participating in data annotation. Fifty-six volunteer medical students and radiology trainees, hereafter referred to as annotation coordinators, were recruited to perform preliminary annotation of data, followed by guided review and refinement by board-certified neuroradiology faculty annotators to establish data 'ground truth' or 'reference standard'. Panel discussions were held with faculty annotators and annotation coordinators focusing on MRI neuroanatomy, pathology, and tumor labeling. Social media pages were set up (LinkedIn, Instagram, X) to post educational material on imaging features of tumors and interesting cases to highlight this initiative to a broad audience. To better understand the status of AI-focused education in medical school and the need for learning opportunities, 54 medical students from the US and across Europe were surveyed with the questionnaire in Supplementary Table 1.

BraTS Lighthouse Challenge 2025:

For the MICCAI BraTS Lighthouse Challenge 2025, a unique approach was adopted to establish the data reference standard. Board-certified neuroradiology faculty annotators with over 10 years of experience were paired with annotation coordinators one-on-one to annotate images in 4 stages, with each subsequent stage separated by a 7-day interval to suppress image memory and recall - (i) Manual annotation from scratch (ii) Repeated manual annotation from scratch (iii) Review and refinement of automated nnU-Net pre-segmented images (iv) Repeated review and refinement of automated nnU-Net pre-segmented images (Figure 1).

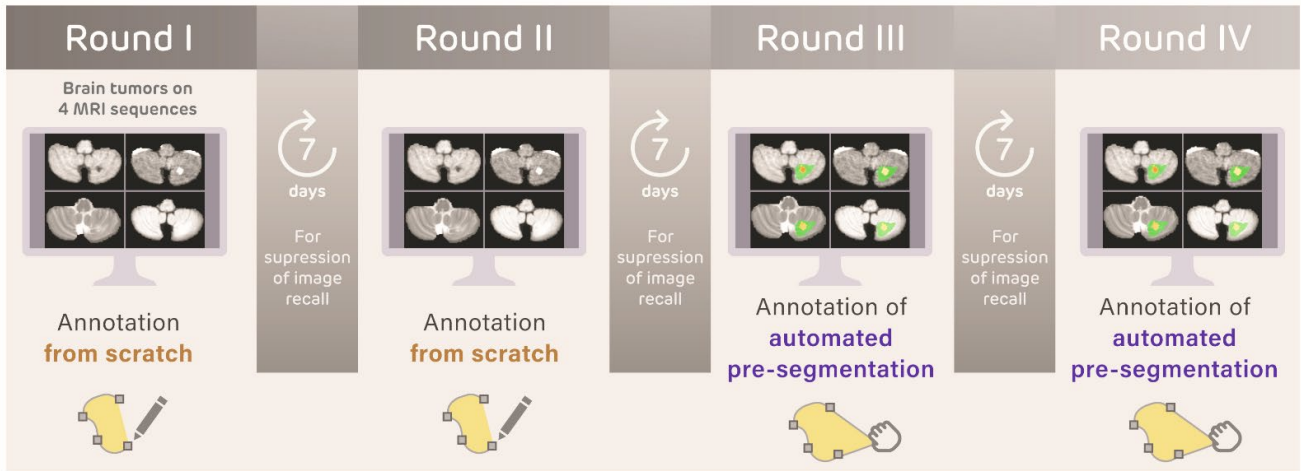

Figure 1. Schematic illustration of the segmentation pipeline of the BraTS Lighthouse Challenge 2025. (Figure prepared by B. Bennett. CMI @ 2025 Children's Hospital of Philadelphia. All rights reserved. Used with permission.)

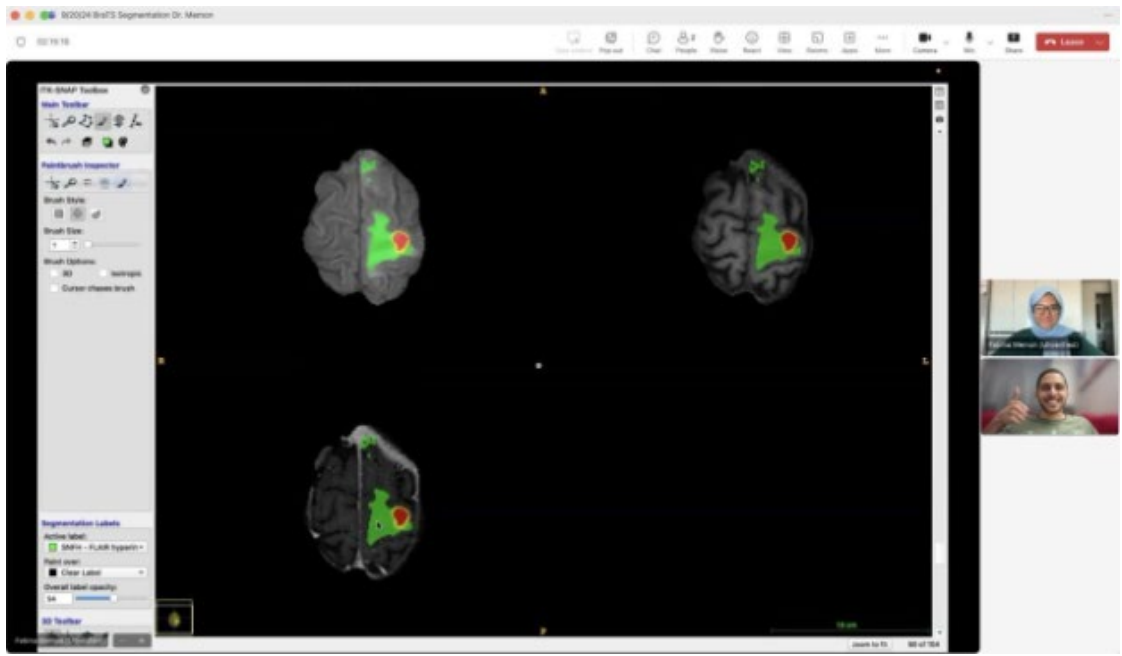

Figure 2. A virtual one-on-one guided annotation session with a neuroradiology faculty annotator and annotation coordinator using ITK-SNAP software to annotate brain metastases

Segmentation Software:

Image annotation was performed using ITK SNAP (Figure 2) (version 4.2.0; https://www.itksnap.org/pmwiki/pmwiki.php?n=Downloads.SNAP4) and 3D slicer (version 5.6.2; https://download.slicer.org/)[19]. Annotation coordinators were guided in using ITK-SNAP through one-on-one meetings with organizers, video tutorials, and documents with best practice guidelines to identify and label brain tumor sub-regions (https://www.youtube.com/watch?v=tvI5BiMs8C4). Two annotation coordinators were paired with faculty experienced in operating 3D Slicer. Coordinators were surveyed with questions from Supplementary Table 2 before and after the annotation pipeline.

Online education platform:

To highlight this initiative to a broad audience of learners globally, several educational events including lectures, workshops, and journal clubs were organized and promoted by leveraging social media platforms like LinkedIn, X, Instagram, BlueSky, YouTube, and WhatsApp.

Lectures:

Neuroradiology faculty-led lectures on fundamentals of MRI, neuroanatomy, brain tumors, and artificial intelligence were organized and highlighted through social media, hosted live and later made available on YouTube for asynchronous learning (https://www.youtube.com/@BraTS_METS_2025_Lighthouse/videos). Attendees completed surveys on their perceived knowledge levels before and after the lectures and provided qualitative feedback on the sessions (Supplementary Table 3).

Bite-sized learning:

Recognizing time constraints during medical school and radiology training, bite-sized information on neuroanatomy, brain tumor features, sub-regions, and interesting cases was posted on our social media platforms for concise, time-friendly learning.

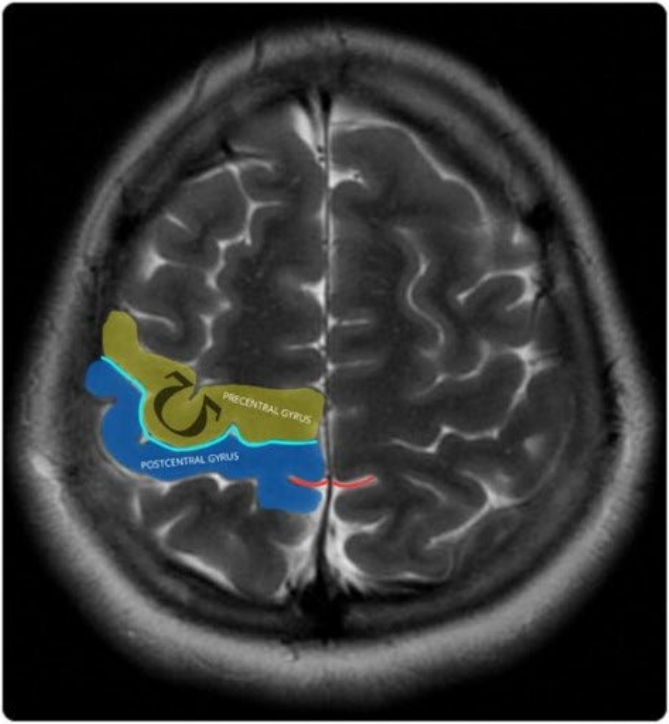
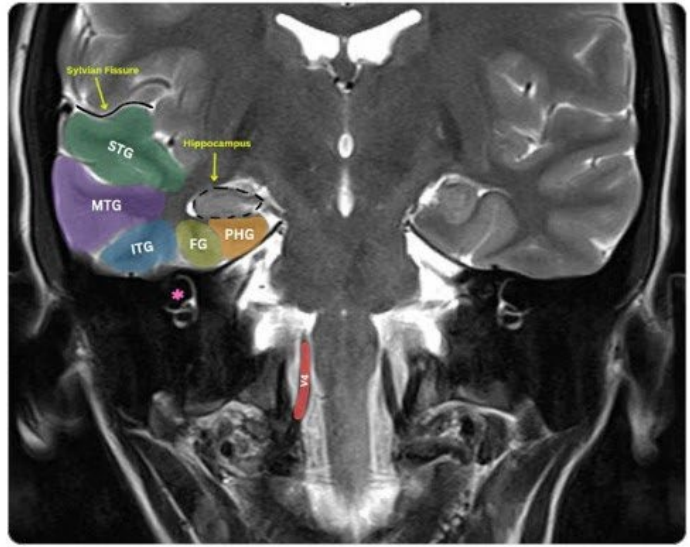

Figure 3. Examples of educational posts on social media showing features of normal anatomy on MRI (https://bsky.app/profile/brats2025.bsky.social/post/3lek7faqba22e; https://bsky.app/profile/brats2025.bsky.social/post/3lgwhvselec2f)

Workshops:

Workshops with hands-on engagement and problem solving led by data scientists were hosted to introduce beginners to the stages of the BraTS Challenge, fundamentals of algorithm development, and recommendations for improving accuracy while training algorithms (Figure 4).

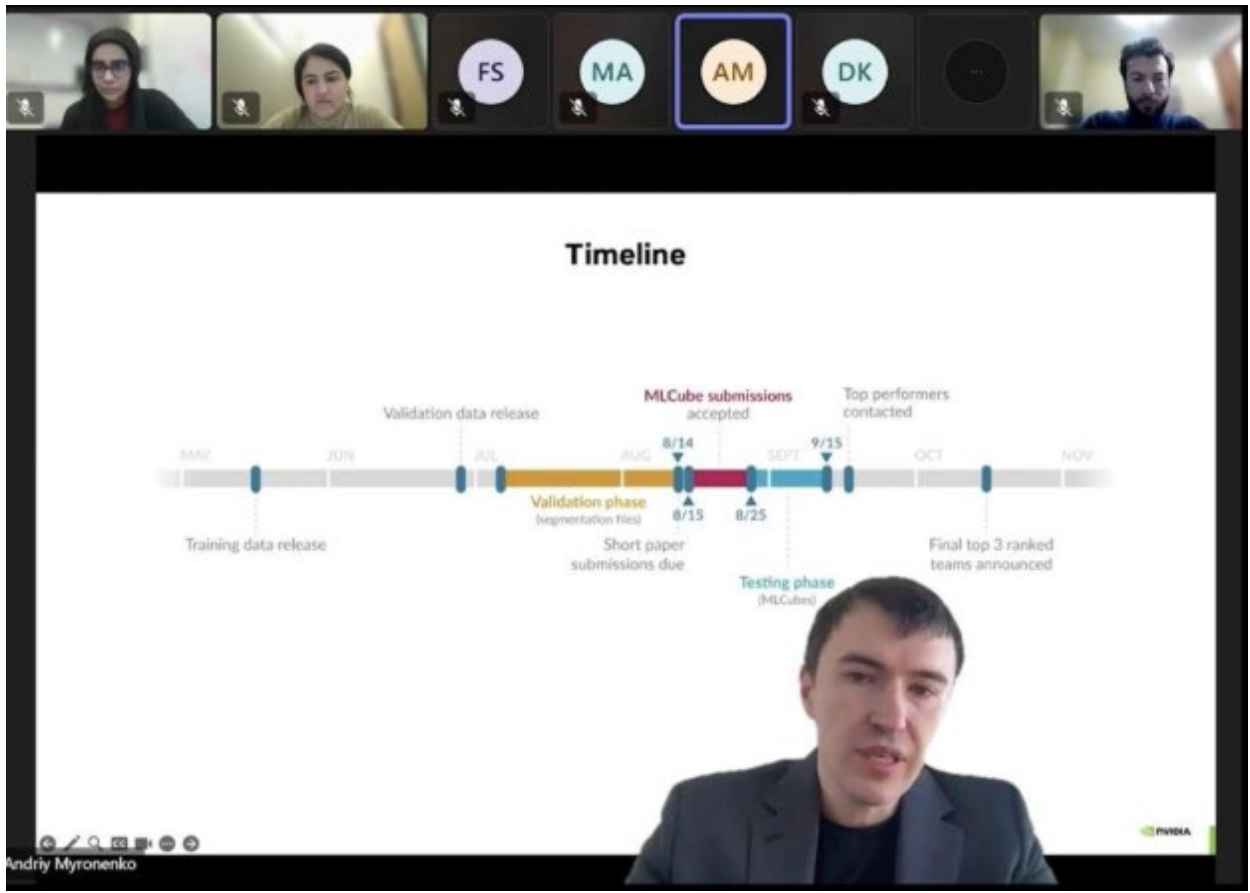

Figure 4. Example of a workshop led by an NVIDIA data scientist explaining the timeline of the BraTS Challenge on video conference.

Journal Clubs:

Student- and trainee-led journal clubs were hosted involving an initial article summary to an audience of peers, an invited panel of authors and subject matter experts. The inclusion of authors and experts allowed for a subsequent robust panel discussion.

The software tutorial, lectures, workshops, and journal clubs uploaded to YouTube can be accessed at https://www.youtube.com/@BraTS_METS_2025_Lighthouse/videos. An overview of educational initiatives offered through the BraTS Lighthouse Challenge is depicted in Figure 5.

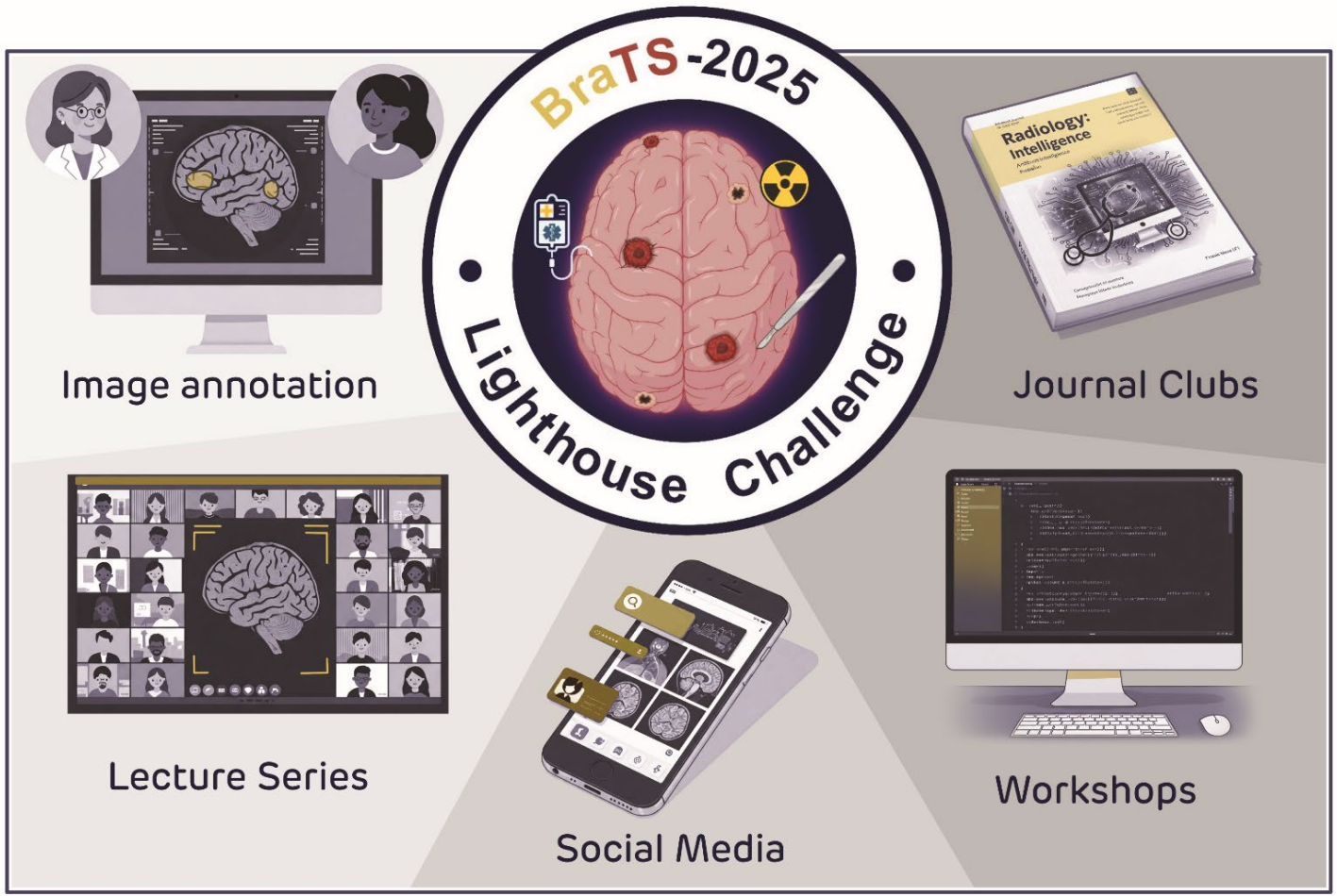

Figure 5. An overview of educational initiatives offered through the BraTS Lighthouse Challenge 2025. (Figure prepared by B. Bennett. CMI @ 2025 Children's Hospital of Philadelphia. All rights reserved. Used with permission.)

**RESULTS**

BraTS Challenges of 2023 and 2024:

To assess the need for an educational chapter within the BraTS Challenge, at the end of the data annotation pipeline for the 2023 edition of the Challenge, fifty-four medical students were surveyed using the questionnaire in Supplementary Table 1 and their responses are depicted in Figure 6. Among them, 13% (7/54) were in the first and second (preclinical) years of medical school, 52% (28/54) were in the third and fourth (clinical) years of medical school, and 33% (18/54) were interns or research fellows.

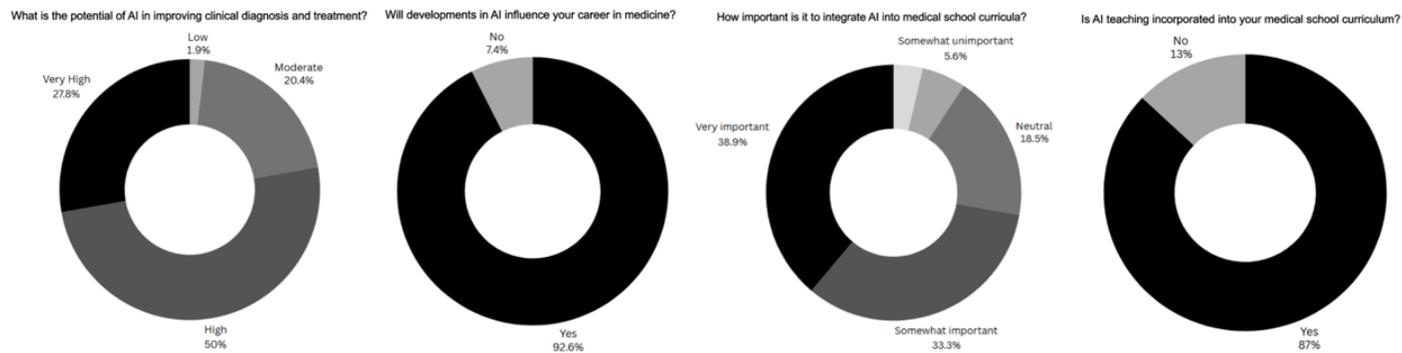

Figure 6. Donut charts depicting responses to the questionnaire in Supplementary Table 1 from 54 annotation coordinators

BraTS Lighthouse Challenge 2025

Annotation pipeline:

During the BraTS Lighthouse Challenge 2025, 14 annotation coordinators were recruited to participate in guided one-on-one data annotation sessions with 14 board-certified neuroradiology faculty annotators. Each faculty-coordinator pair spent an average of 1322.9 ± 760.7 hours completing the 4-stage annotation process to establish the reference standard for each dataset. 6 pairs annotated brain metastases, 2 pairs annotated gliomas, 3 pairs annotated untreated meningiomas, and 3 pairs annotated post-treatment meningioma images. A total of 1200 image annotations were completed during this pipeline, thus contributing valuable reference standard data to the scientific community. Figure 7 shows box plots for the time spent per case for each tumor segmentation task. The higher median time and the wider interquartile range for the time spent by glioma annotators demonstrate that glioma MRI cases consistently require more time to be annotated due to high overall case complexity.

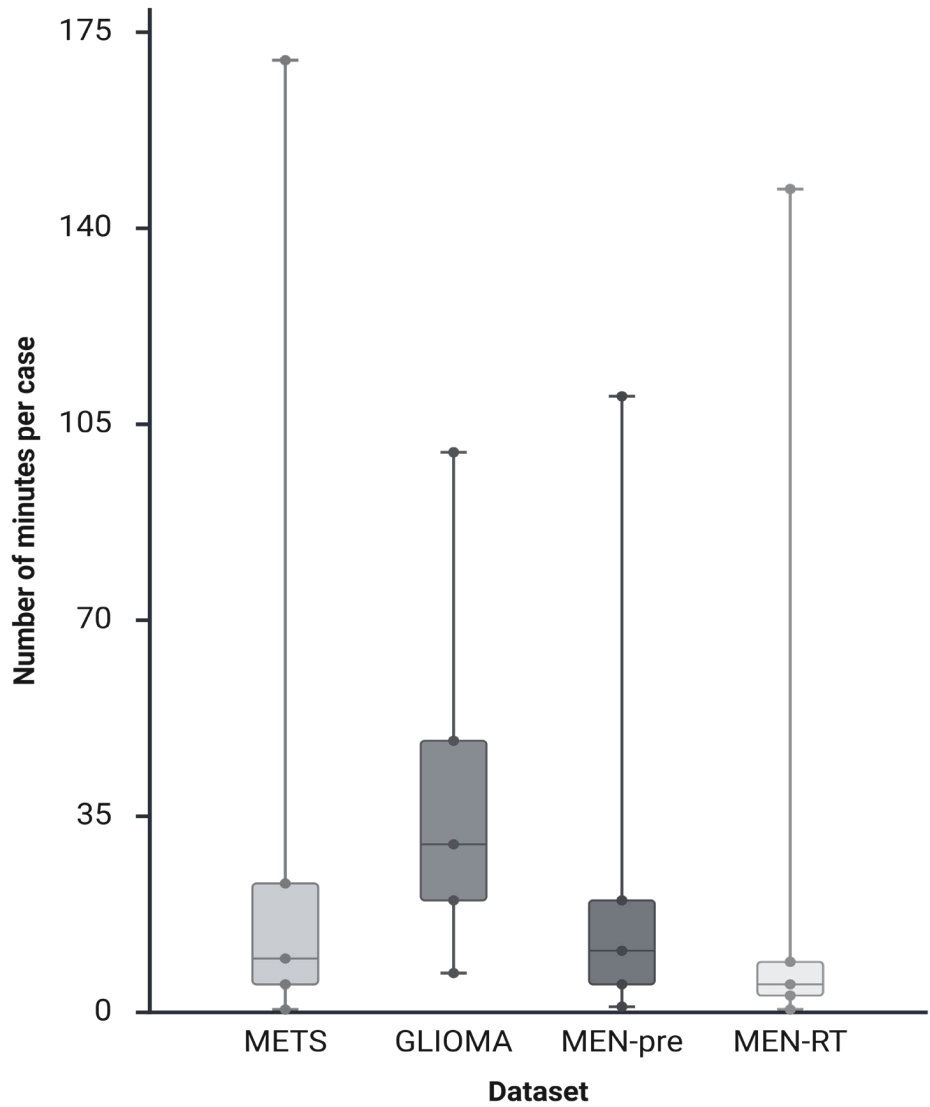

Figure 7. Box plots summarizing the amount of the time spent per case for each tumor segmentation task

Note: METS = metastases; GLIOMA = pre- and post-treatment gliomas; MEN-pre = pre-treatment meningioma; MEN-RT = post-radiation therapy meningioma

Four rounds of annotation for the same dataset were performed independently by 2 faculty-coordinator pairs to assess intra- and inter-annotator variability in tumor labeling, which plays a crucial role in the accuracy of algorithms trained using the data. The sessions also provided trainees with a valuable opportunity to develop mentor-mentee relationships with faculty annotators, simulating an internship in brain tumor imaging, nuances of MRI, quality control, and teamwork in AI-assisted medical imaging.

The annotation coordinators completed the same survey (Supplementary Table 2) to assess their perceived knowledge before and after the guided one-on-one annotation sessions. Their ratings on a scale of 1-10, with 1 being the lowest, and 10 being the highest, are represented in Figure 8. While their average baseline familiarity with image segmentation tools was rated 6 ± 2.93, an increase in average familiarity to 8.9 ± 1.07 was noted at the end of the annotation pipeline, demonstrating the perceived effectiveness of interactive learning (Student's t test p<0.05). While their average baseline familiarity with brain tumor imaging was rated 6.2 ± 2.4, an increase in average familiarity to 8.1 ± 1.2 was reported at the end of the annotation pipeline, demonstrating the perceived effectiveness of one-on-one mentorship (Student's t test p<0.05).

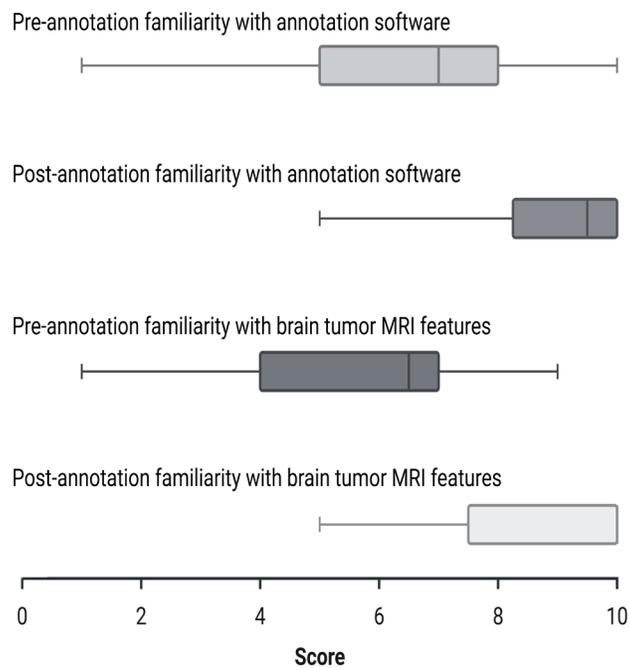

Figure 8. Plots of self-reported scores of 14 annotation coordinators on pre- and post-annotation familiarity levels of image segmentation software and brain tumor MR imaging features

Lectures:

Faculty-led lectures included topics listed in Table 1. Some videos from our educational sessions were not uploaded to YouTube out of consideration for faculty members' preferences.

Table 1. List of lectures hosted during the BraTS 2025 Lighthouse Challenge

| List of lectures hosted during the BraTS Lighthouse Challenge 2025 |
| --- |
| Clinically relevant MRI neuroanatomy |
| Fundamentals of magnetic resonance imaging |
| Standardized reporting for brain tumors |
| Molecular imaging of brain tumors |
| Pediatric brain tumors |
| Skull base imaging |
| Fundamentals of Artificial Intelligence |
| Artificial Intelligence in brain tumor imaging |
| Implementing artificial intelligence in low-resource settings |

A total of 97 lecture attendees responded to questions in Supplementary Table 3 before and after the lectures. Figure 9 shows that attendees reported diverse levels of perceived knowledge before attending lectures, with the majority (50%, n=48) reporting it as "moderate". On the post-lecture survey, a total of 95% (n=92) agreed that their perceived knowledge of the topic had improved, with 6% reporting as "somewhat agree", 29% reporting as "agree", and 60% reporting as "strongly agree". Additionally, qualitative feedback from the attendees at the end of each lecture provided valuable insights for future directions of improvement.

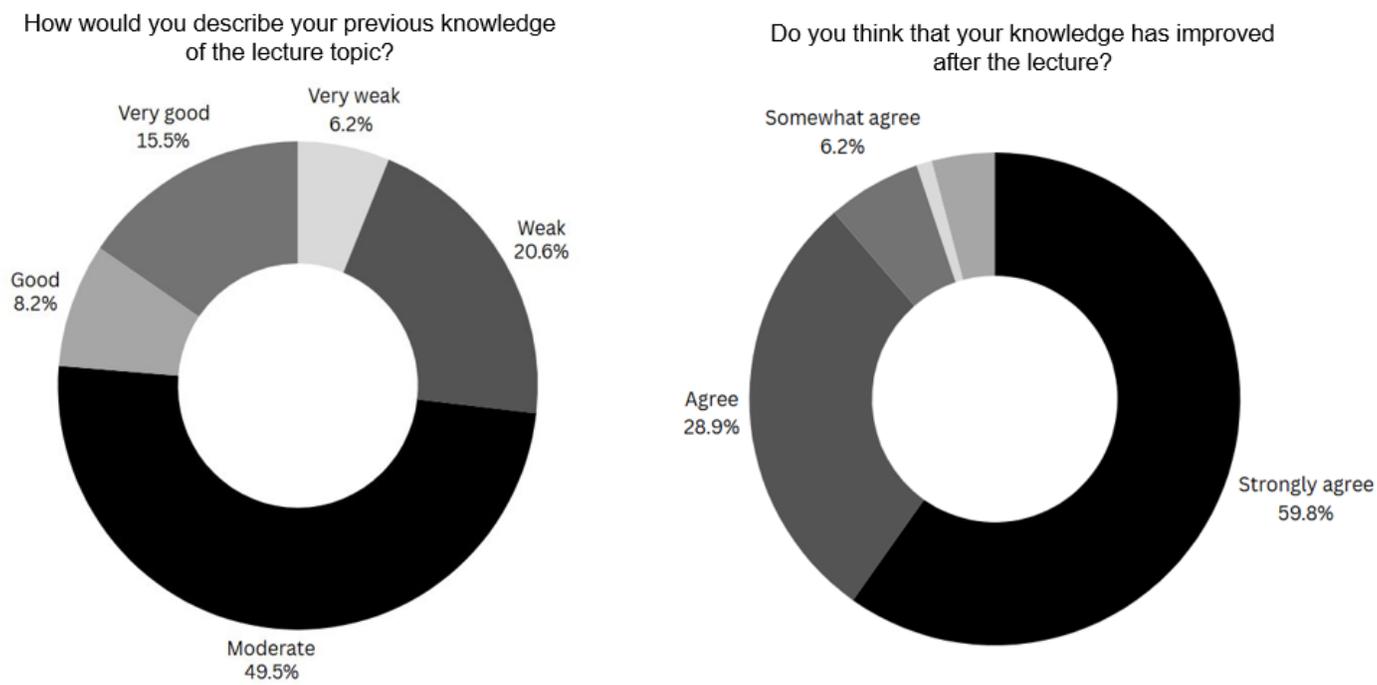

Figure 9. Responses to pre- and post-lecture questions from Supplementary Table 3 for 97 attendees' perceived knowledge

Workshops:

The workshops led by data scientists hosted in 2025 addressed topics including practical strategies and approaches to mastering image segmentation challenges, exploring and setting up virtual environments for algorithm training, and creating an AI-based image segmentation model from scratch. This resulted in 2 submissions of algorithms to BraTS 2025 Lighthouse Challenge by trainees.

Journal Clubs:

The journal clubs hosted in 2025 focused on the following topics: environmental sustainability and AI in radiology, cybersecurity and large language models in radiology, and artificial intelligence in global radiology.

**DISCUSSION**

The educational platform created through the ASNR MICCAI BraTS Lighthouse Challenge 2025 represents an innovative approach toward integration of AI-focused skills into radiology education by linking development of high quality annotated datasets by neuroradiologists and providing a unique opportunity for medical students and radiology trainees to build their knowledge and skills in basics of neuroradiology, brain tumor imaging, and fundamentals of AI. We show that generation of high-quality reference standard data by neuroradiologists is a very time-consuming process that can be augmented by involving medical students and radiology trainees, thus generating an educational benefit. In ASNR MICCAI BraTS 2025, our team of organizers integrated multi-modal educational approaches to prepare annotation coordinators to work with neuroradiology faculty annotators across the globe to generate high quality reference standard data. This was achieved by combining traditional methods of education such as didactic lectures with interactive, practical experiences alongside experts like one-on-one annotation sessions, workshops, and journal clubs. This initiative emphasized the importance of fostering both technical expertise in AI tools and a strong understanding of radiological principles that are crucial to modern healthcare. This initiative has demonstrated the potential to bridge the gap between theoretical knowledge and clinical application by applying clinical knowledge of brain tumor imaging to train and benchmark algorithms with the highest possible accuracy.

The feedback obtained from trainees underscores the importance of embedding AI-focused knowledge into medical training to prepare future radiologists for an increasingly data-driven and AI-supported clinical environment. Results from pre- and post-annotation surveys demonstrated improvement in trainees' familiarity with image segmentation software and understanding of MRI features of brain tumors, emphasizing the effectiveness of interactive learning modalities. One-on-one annotation sessions provided annotation coordinators with an opportunity to directly apply newly learned concepts, gain insights from experienced faculty members, and address knowledge gaps in a collaborative setting. This approach not only improves knowledge retention but also cultivates critical thinking and analytical skills necessary for leveraging AI in clinical practice.

The impact of the education platform built through the ASNR MICCAI BraTS 2025 Lighthouse Challenge highlights the need for transformation in radiology education by incorporating AI as a core component of curricula. Drawing upon the insights gained from this initiative, we advocate for the creation and implementation of a focused, high-yield AI curriculum that would distill the most impactful educational components from the lectures, workshops, and annotation sessions into a structured program within existing medical school and residency schedules. Our experience suggests that best practices include interactive lectures, workshops, and real-world application scenarios emphasizing technical skills in AI-driven healthcare. Such an approach allows consistent engagement with AI-focused content and systematically build trainee knowledge and skills. We also propose a formal framework wherein trainee engagement contributes to the ongoing expansion of robust, expertly curated datasets. By integrating data generation with education, trainees would gain valuable practical experience and actively enrich the repository of available high-quality data, while advancing radiology as a leader in AI. Moving forward, similar initiatives should also focus on promoting interdisciplinary collaboration, addressing ethical considerations in AI integration, and ensuring access to training resources. This initiative also underscores the importance of continuous feedback mechanisms, allowing educators to refine strategies and address evolving needs.

Traditional radiology education places inadequate emphasis on data analysis and applications of AI in image interpretation and clinical workflows. [20] Competency in interdisciplinary collaboration is partly addressed via radiology-pathology correlation requirements for residency graduation, although working effectively with data scientists, engineers, and other specialists – is often underexplored, despite being essential for advancing research and innovation. Ethical considerations and the ability to critically evaluate biases and limitations of AI are also overlooked, potentially leaving radiologists and trainees insufficiently equipped to navigate the challenges of integrating these technologies into patient care. [21]

Another key area of focus of our initiative is fostering a culture of collaboration between medical educators, AI developers, and clinical experts. [22] This interdisciplinary approach ensures that educational initiatives reflect current advancements in technology and are clinically relevant. In addition, this collaborative approach is a method for developing high integrity annotated data that is critically needed in the data science community for the development of clinically relevant AI algorithms. Building on the impact of the one-on-one annotation sessions and workshops, mentorship programs should be established in medical schools and residency programs where students and trainees can work closely with faculty and industry leaders on real-world radiology-based projects. Furthermore, continuous feedback from students and trainees should be given priority, helping educators refine educational content and strategies to improve knowledge retention and skill development. By adopting these measures, the next generation of radiologists will have available robust AI tools built on high integrity annotated data and be well-equipped to effectively navigate the evolving landscape of AI applications in radiology.

## CONCLUSION

ASNR MICCAI BraTS 2025 Lighthouse challenge provided an opportunity to develop a blueprint of educational offerings for medical students and radiology trainees to build their knowledge and skills in the process of developing high quality annotated imaging datasets for brain tumor segmentation algorithm development. We show that there is a critical need for AI-focused education among medical trainees and how novel personalized approaches to teaching neuroradiology can improve training curricula to equip future radiologists to practice in an AI-assisted clinical environment.

# SUPPLEMENTARY TABLES

Supplementary Table 1: Qualitative assessment questionnaire provided to coordinators during BraTS Challenges 2023 & 2024

| Will developments in AI influence your career? | Yes | | | | |
| --- | --- | --- | --- | --- | --- |
| | No | | | | |
| Is AI teaching incorporated in your curriculum? | Yes | | | | |
| | No | | | | |
| How important is it to integrate AI into your curriculum? | Not important | Somewhat unimportant | Neutral | Somewhat important | Very important |
| | | | | | |
| What is the potential of AI in improving healthcare? | Very low | Low | Moderate | High | Very high |
| | | | | | |

Supplementary Table 2: Self-assessment questionnaire provided to annotation coordinators for pre- and post-annotation feedback

| | 1 | 2 | 3 | 4 | 5 | 6 | 7 | 8 | 9 | 10 |
| --- | --- | --- | --- | --- | --- | --- | --- | --- | --- | --- |
| On a scale of 1-10, with 1 as the lowest and 10 as the highest, rate your familiarity with image segmentation software | | | | | | | | | | |
| On a scale of 1-10, with 1 as the lowest and 10 as the highest, rate your familiarity with MRI features of brain tumors | | | | | | | | | | |

Supplementary Table 3: Qualitative assessment questions provided to lecture attendees for the BraTS Lighthouse Challenge 2025

| **Pre-Lecture** | | | | | |
| --- | --- | --- | --- | --- | --- |
| How would you describe your previous knowledge of the topic? | Very weak | Weak | Moderate | Good | Very good |
| | | | | | |
| **Post-Lecture** | | | | | |
| Do you think that your knowledge has improved after the lecture? | Strongly disagree | Disagree | Somewhat disagree | Agree | Strongly agree |
| | | | | | |